\documentclass[journal]{IEEEtran}
\ifCLASSINFOpdf
\else
\fi
\usepackage{amssymb}
\usepackage{caption}
\usepackage{subcaption}
\usepackage{graphicx} 
\usepackage{pifont}
\usepackage{array}
\usepackage{graphicx}
\usepackage{cite}
\newcommand{\xmark}{\ding{55}}
\usepackage{comment}
\usepackage{amsmath,amssymb} 
\usepackage{color}
\usepackage{multirow}
\usepackage{multicol}
\usepackage{booktabs}
\makeatletter
\let\NAT@parse\undefined
\makeatother
\usepackage{hyperref}  
\usepackage[switch]{lineno}
\begin{document}
%
\title{Towards Lightweight Transformer via Group-wise Transformation for Vision-and-Language Tasks}
%
%
%

\author{Gen~Luo,
        Yiyi~Zhou*,
        Xiaoshuai Sun,~\IEEEmembership{Member,~IEEE,}
        Yan Wang,~\IEEEmembership{Member,~IEEE,}
        Liujuan Cao,
        Yongjian Wu,
        Feiyue Huang,
        Rongrong Ji,~\IEEEmembership{Senior~Member,~IEEE }

        \thanks{G. Luo, Y. Zhou, X. Sun, L. Cao and R. Ji are with the Media Analytics and Computing Lab, Department of Artificial Intelligence, Xiamen University, China e-mail: (luogen@stu.xmu.edu.cn, zhouyiyi@xmu.edu.cn, caoliujuan@xmu.edu.cn, xssun@xmu.edu.cn, rrji@xmu.edu.cn).}
        \thanks{Y. Wang is with
        	Microsoft, China e-mail: (yanwang@ee.columbia.edu
        	).}
        \thanks{Y. Wu and F. Huang are with the Tencent Technology (Shanghai) Co., Ltd, China e-mail: (littlekenwu@tencent.com, garyhuang@tencent.com).}
        \thanks{*Corresponding Author: Yiyi~Zhou (E-mail: zhouyiyi@xmu.edu.cn)}
}

%
%

\markboth{Journal of \LaTeX\ Class Files,~Vol.~14, No.~8, August~2015}%
{Shell \MakeLowercase{\textit{et al.}}: Bare Demo of IEEEtran.cls for IEEE Journals}
%



\maketitle

\begin{abstract}
	Despite the exciting performance, Transformer is criticized for its excessive parameters and computation cost.
However, compressing Transformer remains as an open problem due to its internal  complexity of the layer designs, \emph{i.e.}, \emph{Multi-Head Attention} (MHA) and \emph{Feed-Forward Network} (FFN). 
To address this issue, we introduce  \emph{Group-wise Transformation}  towards  a universal yet lightweight  Transformer for vision-and-language tasks, termed as \emph{LW-Transformer}\footnote{Source codes are released \hyperlink{}{https://github.com/luogen1996/LWTransformer}.}.  
LW-Transformer applies Group-wise Transformation to reduce both the parameters and computations  of Transformer, while also preserving  its two main properties, \emph{i.e.}, the efficient attention modeling on diverse subspaces of MHA, and the expanding-scaling feature transformation of FFN. 
We apply LW-Transformer to a set of Transformer-based networks, and quantitatively measure them on three vision-and-language tasks  and six benchmark datasets.
Experimental results show that while saving a large number of parameters and computations,  LW-Transformer  achieves very competitive performance against the original Transformer networks  for vision-and-language tasks.  To examine the generalization ability, we  also apply our optimization strategy to a recently proposed image Transformer called Swin-Transformer for image classification, where the  effectiveness  can  be  also confirmed. 
\end{abstract}

\begin{IEEEkeywords}
Lightweight Transformer, Visual Question Answering, Image Captioning, Reference Expression Comprehension.
\end{IEEEkeywords}

%
\IEEEpeerreviewmaketitle

\section{Introduction}
%
%
%
%

\begin{figure*}[t]
	\centering
	\includegraphics[width=2\columnwidth]{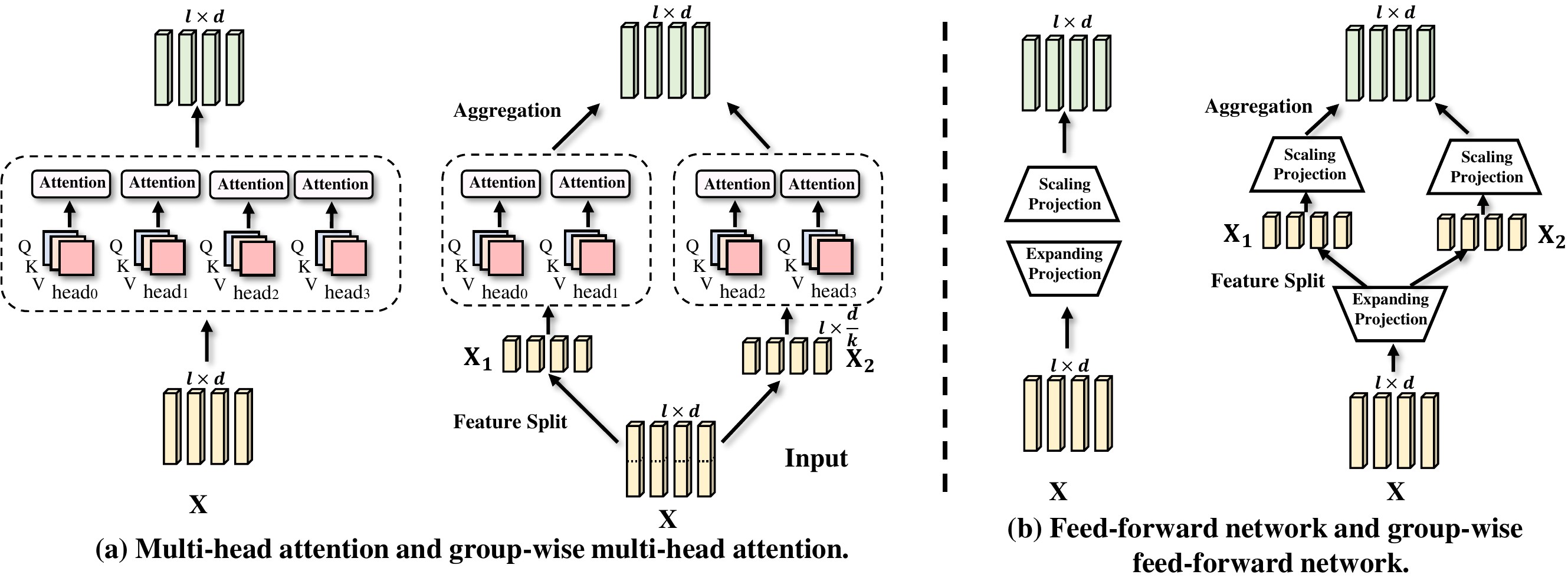}
	
	\caption{ Illustration of Group-wise Multi-Head Attention (G-MHA) and Group-wise Feed-Forward Network (G-FFN). (a) The proposed G-MHA divides input feature  into $k$ groups, of which the multi-head attention is conducted, and the outputs are concatenated to one feature. (b) G-FFN fisrt performs expanding projections and then conducts group-wise linear transformation.}
	\label{MHA}
	\vspace{-4mm}
\end{figure*}
\IEEEPARstart{T}{ransformer}~\cite{vaswani2017attention} has become the model of choice in the field of natural language processing  due to its superior capability of modeling long-range dependencies and learning effective representations.  Its great success has also attracted the attention of the computer vision community~\cite{yu2019deep,huang2019attention,yu2019multimodal,carion2020end}. At present, Transformer and its variants have been applied to a variety of  vision-and-language tasks, such as \emph{visual question answering} (VQA)~\cite{gao2019dynamic,gao2019multi,yu2019deep}, \emph{image captioning}~\cite{cornia2019m,herdade2019image,huang2019attention,li2019entangled}, and \emph{referring expression comprehension}~\cite{yu2019multimodal,yu2020deep},  achieving dominant performance in multiple benchmarks. 
Besides, Transformer also yields a new trend of large-scale multi-modal pretraining~\cite{alberti2019fusion,devlin2018bert,li2019unicoder,li2019visualbert,lu2019vilbert,lu201912,su2019vl,tan2019lxmert,yang2019xlnet,zhou2019unified}, which greatly promotes the development of joint vision-and-language learning.

Coming with the outstanding  ability of long-distance dependency modeling, Transformer-based models are also  known for their extreme demand on computation  power  and memory space. Such an issue will become more prominent when applying Transformer to vision-and-language tasks, where the multi-modal network is typically built based on a large visual backbone, \emph{e.g.}, ResNet~\cite{he2016deep} or Faster R-CNN~\cite{ren2015faster}.
For instance, VL-BERT-large~\cite{su2019vl} needs to be trained  on 8 GPUs in parallel for nearly 20 days, and its parameters are about 340 millions, which takes about 1.2 GB of memory  to deploy.  Thus, the large latency and memory footprint greatly hinder the application of Transformer on the mobile devices. 

To this end, the network compression for
Transformer-based models has attracted significant research attention~\cite{fan2019reducing,sanh2019distilbert,shen2019q,tian2019waldorf,zafrir2019q8bert}. 
Popular directions include \textit{distillation}~\cite{sanh2019distilbert,tian2019waldorf}, \textit{layer pruning}~\cite{fan2019reducing} and \textit{quantization}~\cite{shen2019q,zafrir2019q8bert}.
Beyond these methodologies, one  orthogonal direction is to directly optimize the Transformer layer itself~\cite{ma2019tensorized,choromanski2020rethinking,wang2020linformer}, which has not been paid enough attention and  is the focus of this paper.
By designing a more efficient version of Transformer that still preserves the accuracy, one can  directly train a model based on the improved structure to   avoid the  training-distilling process~\cite{kitaev2019reformer}, which can be also  combined with the popular compression methods to further improve their efficiency.

However, designing an efficient Transformer remains as a  challenging  problem, mainly due to the unique structures of the Multi-Head Attention (MHA) and the Feed-Forward Network (FFN). First, as a key component in Transformer, MHA is used to capture the dependency of input features in various subspaces, upon which efficient feature transformations can be obtained~\cite{vaswani2017attention}. In this case, these transformations in MHA not only  map features to new semantic   spaces but also   compute the attention scores using the \emph{softmax} function. 
Second, as a layer-to-layer connection design, FFN achieves efficient transformations by expanding and scaling the feature dimensions, which inevitably involves a large number of parameters. Both designs pose a huge obstacle in reducing Transformer  parameters.

In this paper, we  introduce \emph{Group-wise Transformation} to achieve a  lightweight Transformer  for vision-and-language tasks, termed \emph{LW-Transformer}, which well addresses the aforementioned issues  in a principled way. 
Group-wise Transformation is an efficient  method used in many convolution neural networks (CNNs)~\cite{chollet2017xception,krizhevsky2012imagenet,sandler2018mobilenetv2,szegedy2017inception,szegedy2016rethinking,zhang2018shufflenet}. It  splits the input features  into $k$ groups by channels to perform  parameterized transformations, and then  concatenates the transformed features as the output. This  group-wise operation can reduce the  parameters and computations  of the original transformation  by $\frac{k-1}{k}$.  Based on this strategy, we propose the \emph{group-wise multi-head attention} (G-MHA) and \emph{group-wise feed forward networks} (G-FFN) for LW-Transformer.

Notably,  in addition to improve model compactness, LW-Transformer  can well maintain the desired principles of  the original  Transformer.   { The effectiveness of  Transformer  mainly lies in its  MHA design~\cite{vaswani2017attention}, which captures  the diverse  feature dependencies  through  different  feature  subspaces.}
These subspaces are obtained by truncating the input features after linear projections, as shown in Fig.~\ref{MHA}. 
The principle of Group-wise transformation is essentially consistent with that of MHA, which also aims to  learn more efficient feature representations  through  more diverse feature spaces. 
In this case, the proposed G-MHA takes a step forward than MHA, which directly splits input features before projections. More importantly, this operation does not change neither the feature dimension nor the scaled-dot product for attention modeling. Meanwhile,  G-FFN   also maintains the expanding-scaling property of FFN while reducing the computation and parameters to a large extent, as shown in Fig.~\ref{MHA}.

 During experiments, we build LW-Transformer based on the structures of the default Transformer and its variants, including  MCAN \cite{yu2019deep}, LXMERT \cite{tan2019lxmert}  and Swin-Transformer~\cite{liu2021swin}, and conduct extensive experiments on six benchmark datasets,~\emph{i.e.,} GQA~\cite{hudson2019gqa}, VQA2.0~\cite{goyal2017making}, CLEVR~\cite{johnson2017clevr}, COCO~\cite{chen2015microsoft}, RefCOCO~\cite{yu2016modeling}, RefCOCO+~\cite{yu2016modeling}, of three language-and-vision tasks, \emph{i.e.,} VQA~\cite{goyal2017making}, image captioning~\cite{chen2015microsoft} and referring expression comprehension (REC)~\cite{kazemzadeh2014referitgame}.  We also test LW-Transformer on the large-scale and BERT-style pretraining~\cite{tan2019lxmert}  and the task of image classification.
The experimental results show that with an overall reduction of up to 45\% of parameters and 28\% of computations\footnote{Under the setting of LWTransformer$_{1\times}$.}
, LW-Transformer  still achieves  competitive performance against the default Transformer on six benchmark datasets of 
  the three  VL tasks and even obtains performance gains on some benchmarks, \emph{e.g.,}  +1\% on GQA, +0.1\% on COCO Captioning and +3.2\% on RefCOCO, which well confirms  its effectiveness  for vision-and-language tasks. 
  In addition, its generalization ability is also well validated on ImageNet and CIFAR-100. 
  Compared with Swin-Transformer, LW-Transformer can obtain the same performance on CIFAR-100 while saving 42.4\% computations and 43.3\% parameters. On ImageNet, LW-Transformer can even obtain better performance with similar experimental cost to Swin-Transformer. 

As a summary, our contributions  are three-fold:
\begin{itemize}
	\item We  present the first application of  Group-wise transformation to Transformer, and propose an efficient Transformer network called LW-Transformer. 
	\item On six benchmarks of three language-and-vision tasks, LW-Transoformer  achieves competitive performance against the default Transformers with a reduction of up to 45\% parameters and 28\% computations. Its generalization  ability  is also validated on  the recently proposed  Swin-Transformer~\cite{liu2021swin}  for image classification.  
	\item We conduct extensive experiments to examine the optimizations of different designs in Transformer, which can provide useful insights for its future optimization.
\end{itemize}
\section{Related Work}
Due to the superior capability of learning efficient representations, Transformer~\cite{vaswani2017attention} is becoming  popular  in both natural  language processing \cite{ma2019tensorized,so2019evolved,sukhbaatar2019adaptive,vaswani2017attention} and computer vision~\cite{cornia2019m,gao2019dynamic,gao2019multi,herdade2019image,huang2019attention,li2019entangled,yu2019deep}.
A set of  Transformer-based approaches have been proposed to achieve  the state-of-the-art  performance  in VQA~\cite{gao2019dynamic,gao2019multi,yu2019deep,zhou2020k} and image captioning~\cite{herdade2019image,huang2019attention,li2019entangled,pan2020x,liu2018show,liu2017improved,yao2019hierarchy,ji2021improving}. For example, the recent X-Linear Transformer~\cite{pan2020x}  integrates bilinear pooling~\cite{fukui2016multimodal} and squeeze-excitation module~\cite{hu2018squeeze} into self-attention to selectively capitalize on visual information and to perform multimodal reasoning. With these novel  designs, X-Linear Transformer  improves the SOTA performance of image captioning on the CIDEr~\cite{vedantam2015cider} metric by  +1.6.  Meanwhile, these Transformer-based networks also lead to a trend of large-scale language-and-vision pre-training~\cite{li2019unicoder,li2019visualbert,lu2019vilbert,lu201912,su2019vl,tan2019lxmert,zhou2019unified,li2021scheduled}, which break the best performance  of  various tasks, such as Visual Question Answering~\cite{antol2015vqa} and Image Captioning~\cite{chen2015microsoft}.
In addition,  a lot of vision transformers~\cite{liu2021swin,el2021xcit} have been proposed recently, which shows the superior performance on image classification.  For example, XCiT~\cite{el2021xcit} achieves 86.0 \% top-1 accuracy on ImageNet,  outperforming conventional CNNs by a large margin.
Despite the great success, Transformer and its variants have been criticized for the excessive parameters and high computation overhead.  For instance, the parameter sizes of  BERT~\cite{devlin2018bert}, ViLBERT~\cite{lu2019vilbert}  and LXMERT~\cite{tan2019lxmert} are 85, 221 and 181 millions, respectively. 
Such a large amount of parameters greatly hinder the applications of  Transformer and Transformer-based models on the mobile  applications, where both the storage and computation resources are limited.

{Several recent advances~\cite{ma2019tensorized,choromanski2020rethinking,wang2020linformer} are committed to  design the  efficient Transformer networks. Specifically, Linformer~\cite{wang2020linformer} reduces the computation cost of self-attention by compressing the sequence length of input features. Performer~\cite{choromanski2020rethinking} applies the scalable kernel methods to approximate the scale-dot product attention to reduce computation overhead when  the  sequence length is very large. Tensorized Transformer~\cite{ma2019tensorized} uses tensor decomposition to reduce the parameters of Q, K and V in self-attention. Compared with these efficient Transformers, the proposed LW-Transformer neither changes the length of input sequences nor the definition of attention modeling. }

The principle of the introduced Group-wise Transformation is also related to the \emph{group convolution}, which  can be traced back to a set of classical  convolutional  neural networks~\cite{chollet2017xception,howard2017mobilenets,krizhevsky2012imagenet,sandler2018mobilenetv2,szegedy2017inception,szegedy2016rethinking,zhang2018shufflenet} like AlexNet~\cite{krizhevsky2012imagenet}.  The depth-wise separable convolution in Xception~\cite{chollet2017xception} generalized the  method into Inception  Networks~\cite{szegedy2017inception,szegedy2016rethinking}.  ShuffleNets~\cite{zhang2018shufflenet} and MobileNets~\cite{howard2017mobilenets,sandler2018mobilenetv2} use  the group-wise and depth-wise separable convolution to achieve  lightweight deep networks with  competitive performance. However, due to the huge difference between the principles of multi-head self-attention (MHA) and convolution, its application to Transformer networks  is still left unexploited. Except for improving model efficiency, the motivation of this paper is  different from the mentioned CNNs. Existing methods like Xception~\cite{chollet2017xception} use group-wise transformation to realize the multi-branch structure, so as to increase the capacity of the model. In contrast, our application is to inherit the property of global dependency in self-attention, as discussed in the introduction.

\section{Approach}
In this section, we first recap the main components of Transformer,   introduce Group-wise Transformation, and then  describe the proposed LW-Transformer  in detail.
\subsection{Preliminary} \label{prel}
 In Transformer, each layer  typically contains two  main components, \emph{i.e.}, Multi-Head Attentions (MHA) for dependency modeling and Feed-Forward Network (FFN) for position-wise transformations.

\noindent \textbf{Multi-Head Attention}  is used to capture the dependency between input features by the \emph{scaled dot-product attention}~\cite{vaswani2017attention}, upon which efficient representations are learned. 
Specifically, given  the transformed  matrices of \emph{query} $\mathbf{Q}\in\mathbb{R}^{n\times d}$, \emph{key} $\mathbf{K}\in\mathbb{R}^{n\times d},$ and \emph{value} $\mathbf{V}\in\mathbb{R}^{n\times d}$, the scaled dot-product attention function is formulated as:

\begin{equation}
	\begin{aligned}
		\text{Attention}(\mathbf{Q},\mathbf{K},\mathbf{V})=\text{Softmax}(\frac{\mathbf{K}^T \mathbf{Q}}{\sqrt{d}})\mathbf{V}.
	\end{aligned}
	\label{att_func}
\end{equation}
Here, $n$ and $d$ denote the sequential length and the feature depth, respectively.
To further improve the representational ability, Vaswani \emph{et al.}~\cite{vaswani2017attention}  extends Eq. \ref{att_func} to a multi-head version,  denoted as \emph{multi-head attention} (MHA),  which consists $h$ paralleled  ``heads'' to make the scale dot-product attention. Specifically,  given the input features $\mathbf{X} \in\mathbb{R}^{n\times d}$, MHA is  formulated as:
\begin{equation}
	\begin{aligned}
		\text{MHA}(\mathbf{X})&=[head_1,...,head_h],\\
		\text{where}~ head_i&= \text{Attention}(\mathbf{Q}^i,\mathbf{K}^i,\mathbf{V}^i).
		\label{mha_func}
	\end{aligned}
\end{equation}
Here,  $\mathbf{Q}^i$, $\mathbf{K}^i$ and $\mathbf{V}^i$ are the split features for the i-th head, and they are obtained by truncating the input features X after projection.  Meanwhile, a linear-merge  projection is  used after  MHA, defined as:
\begin{equation}
	\begin{aligned}
		\mathbf{O}&=\text{MHA}(\mathbf{X})\mathbf{W}_O,
	\end{aligned}
\end{equation}
where $ \mathbf{W}_O\in \mathbb{R}^{d \times d}$ is the weight matrix  and $\mathbf{O}\in \mathbb{R}^{n \times d}$ is the final output.

\noindent \textbf{Feed-Forward Network} makes a position-wise and dense transformation from the input features to the output ones, which keeps the non-linearity by expanding the hidden units in  the \emph{ReLU} layer, as shown in Fig.~\ref{MHA} (c).
This design is beneficial to prevent manifold collapse  caused by \emph{ReLU}, as well as to extend the model capacity~\cite{sandler2018mobilenetv2}.  The formulation of FFN can be written as 
\begin{equation}
	\begin{aligned}
		\text{FFN}(\mathbf{X})&=  \text{max}(0, \mathbf{X}\mathbf{W}_1+ b_1)\mathbf{W}_2 + b_2.
	\end{aligned}
\end{equation}
Here $\textbf{W}_1 \in \mathbb{R}^{d \times d_{f}}$, $\textbf{W}_2 \in \mathbb{R}^{d_{f} \times d}$, $b_1 \in \mathbb{R}^{d_{f}}$ and $b_2 \in \mathbb{R}^{d}$ are weights and bias, and $\text{max}(\cdot)$ is the \emph{ReLU} activation. Typically, $d_f$ is a quadruple as large as $d$.  

\subsection{Lightweight Transformer}
To achieve the lightweight Transformer, we optimize its two main designs, \emph{i.e.}, MHA and FFN, with  \emph{Group-wise transformation}. In this subsection, we begin with the introduction of Group-wise transformation, and then describe its deployment on MHA and FFN in detail.

\noindent \textbf{Group-wise transformation.} 	As shown in Fig.~\ref{MHA}, given the input features $\mathbf{X} \in \mathbb{R}^{n \times d}$, we first  divide it into $k$ groups. Those features of different  groups are transformed by the function $\tau(\cdot)$ and then concatenated again as one complete output. The formulation of  Group-wise Transformation, denoted as $\tau_{Group}\left(\cdot\right)$, is defined by
\begin{equation}
	\begin{aligned}
		\tau_{Group}\left(\mathbf{X}\right)& = [ \mathbf{X}'_1, ..., \mathbf{X}'_k] ,\\
		\text{where}  ~~  \mathbf{X}'_i &= \tau \left( \mathbf{X}_i; \mathbb{\theta} \right).
	\end{aligned}
\end{equation}
Here, $\mathbf{X}_i \in \mathbb{R}^{n \times \frac{d}{k}}$ denotes the truncated  features of the $i$-th group,  $[\cdot]$ denotes concatenation and $\tau \left( \cdot \right)$ can be any parameterized function, \emph{e.g.,} linear/ non-linear  projections or more complex modules like \emph{self-attention}~\cite{vaswani2017attention}. 

With the \emph{split-transform-concatenate} design,  the Group-wise Transformation can save $\frac{k-1}{k}$  of  parameters  and $\frac{k-1}{k}$   of the computations. To further reduce the number of parameters, we  can also share the learnable weights  for each group, consistently resulting in $\frac{k^2-1}{k^2}$   reductions of  parameters.

\noindent \textbf{Group-wise  Multi-Head Attention.}  
	Specifically, we  extend the original Multi-Head Attention (MHA)  in Transformer to a \emph{group-wise} version (G-MHA), as shown in  Fig.~\ref{MHA} (a). 
	
	 As described in Sec.~\ref{prel}, MHA obtains the multiple attentions via truncating Q, K and V after the projection of X. In this case, G-MHA can be considered as its extension in principle, which directly splits the input features before projection.  Concretely, the input features $\mathbf{X} \in \mathbb{R}^{n \times d}$ are first divided into $k$ splits $\mathbf{X}_i \in \mathbb{R}^{n \times d/k}$, based on which the multi-head attentions are performed. Afterwards, those attention outputs are concatenated again.  Then, G-MHA is defined as: 
\begin{equation}
	\begin{aligned}
		\text{G-MHA}(\mathbf{X})&=[\tau(\mathbf{X}_1),...,\tau(\mathbf{X}_k)],\\
		\text{where}~ \tau(\mathbf{X}_i)&= \text{MHA}(\mathbf{X}_i).
	\end{aligned}
\end{equation}

As shown in Fig.~\ref{MHA} (a), {the group operation of G-MHA is conducted on the feature dimension instead of the sequence length. Therefore, each feature can still obtain its coupling coefficients to  the  others through the comparisons on the truncated features, similar to  the default MHA. }  To this end,  the attention patterns learned by G-MHA can be close to that of MHA in principle. 

Notably, we do not replace the linear-merge layer in MHA, \emph{i.e.}, $\mathbf{W}_o$ in Eq.\ref{mha_func}, which is founded to be important for expanding the learning capacity of Transformer and  facilitating the cross-group information exchange.
Overall, G-MHA reduces the parameter size from $4d^2$ to about $2d^2$, and the computational complexity from O$(4nd^2 + n^2d)$ to about O$(nd^2 + n^2d)$.

\noindent \textbf{Group-wise Feed-Forward Network.} We further deploy Group-wise Transformation to FFN, of which implementation is illustrated in Fig.~\ref{MHA} (b). Specifically, given an input $\mathbf{X} \in \mathbb{R}^{n \times d}$, the group-wise feed-forward network (G-FFN)  performs non-linear projections  to expand the feature size by
\begin{equation}
	\begin{aligned}
		\mathbf{H}&= \text{max}(0, \mathbf{X}\mathbf{W}_1+b_1),
	\end{aligned}
\end{equation}
where $\mathbf{W}_1 \in \mathbf{R}^{d \times d_f}$ and  $b_1 \in \mathbb{R}^{d_{f}}$. Then,  the obtained hidden features  $\mathbf{H} \in \mathbb{R}^{n \times d_f}$  are divided into k independent features $\mathbf{H}_i \in \mathbb{R}^{n \times \frac{d_f}{k}}$, which
is further linearly transformed  with sharable weights. G-FFN can be formulated as: 
\begin{equation}
	\begin{aligned}
		\text{G-FFN}(\mathbf{X})&=[\tau(\mathbf{H}_1),...,\tau(\mathbf{H}_k)],\\
		\text{where}~~~ \tau(\mathbf{H}_i)&= \mathbf{H}_i\mathbf{W}_2 + b_2,
	\end{aligned}
\end{equation}
where  $\mathbf{W}_2 \in \mathbf{R}^{\frac{d_f}{k} \times \frac{d}{k}}$ and $b_2 \in \mathbb{R}^{\frac{d}{k}}$. Compared to the original FFN, G-FFN reduces  the parameter size from O($d^2$) to O($\frac{d^2}{k^2}$) and the complexity from O($nd^2$) to O($\frac{nd^2}{k}$).  

As shown in Eq.~8 and Fig.~\ref{MHA} (b),  G-FFN can still maintains the expanding-scaling property of the default FFN. Notably, the group-wise operation is only deployed on the last linear layer, while the first non-linear projection remains intact.  
To explain,  this  design is beneficial to protect the non-linearity  of FFN and prevent the manifold collapse~\cite{sandler2018mobilenetv2}, therefore avoiding the performance degeneration.

\begin{table}[t]
	\centering
	\caption{
		Notations for different settings of  LW-Transformer.
	}
	\begin{tabular}{ll}
		\hline
		\multicolumn{1}{c}{Notations} & \multicolumn{1}{c}{Description}                     \\ \toprule
		\textit{n}$\times$ (the suffix)                      & The expanding multiples of projection dimensions \\ &in    $\textbf{Q}$ and  $\textbf{K}$.            \\
		\textit{mini} (the suffix)                  &  The mini version illustrated in Tab.III. \\
		k                             & The number of groups in the transformations.        \\
		WS                            & Weight sharing.                                     \\
		G-MHA                         & Group-wise multi-head attention.                    \\
		G-FFN                         & Group-wise feed-forward network.                    \\
		G-LML                         & Group-wise linear-merge layer in MHA.                      \\
		G-IL                          & Group-wise intermediate layer in FFN.                      \\ \bottomrule
	\end{tabular}
	\label{notation}
	\vspace{-1em}
\end{table}

\section{Experiments}
To validate the proposed LW-Transformer, we apply it to a set of Transformer and BERT-style models, and conduct extensive experiments on six benchmark datasets, \emph{i.e.,} VQA2.0~\cite{goyal2017making}, GQA~\cite{hudson2019gqa}, CLEVR~\cite{johnson2017clevr}, MS-COCO~\cite{chen2015microsoft}, RefCOCO~\cite{yu2016modeling}, RefCOCO+~\cite{yu2016modeling}, of three  vision-and-language  tasks, \emph{i.e.}, Visual Question Answering (VQA)~\cite{antol2015vqa}, Image Captioning (IC)~\cite{chen2015microsoft} and Referring Expression Comprehension (REC)~\cite{kazemzadeh2014referitgame}. To examine its generalization ability, we  build LW-Transformer based on the newly proposed  Swin-Transformer~\cite{liu2021swin} for image classification. 
\subsection{Deployed Networks}

\noindent \textbf{Transformer}: For VQA, IC and REC, we use the Transformer networks proposed in ~\cite{vaswani2017attention,yu2019deep} as our baseline model, which all follow a classical encoder-decoder structure~\cite{vaswani2017attention}. In the baseline, we set the $d$, $d_f$, $d_h$ defined in Eq.\ref{att_func} - \ref{mha_func}, as  512, 2,048, 64,  respectively. 
The number of attention heads $h$ is 8 for each Transformer layer, and both the encoder and decoder branches are composed of six Transformer layers.
For simplicity, we denote the baseline network as {\textit{Transformer}}. 
We further replace Transformer with the proposed LW-Transformer layers, and keep the rest designs unchanged. We denote the compressed network as \textit{LW-Transformer}. 
In LW-Transformer, the basic settings of $d$, $d_f$, $d_h$ are the same as   Transformer.
The number of groups is set to 2 by default, and each group has 4 attention heads, so the total number of attentions is kept as 8 as Transformer.

\noindent \textbf{Bert-style Model}: 
The deployed Bert-style model is the recently proposed LXMERT~\cite{tan2019lxmert}, of which structure is slightly different from the conventional Transformer network. It has 9 and 6 Transformer encoder  layers for the language and vision modelings, respectively, and 5 cross-modal Transformer layers\footnote{Cross-modal Transformer layer has two MHA and FFN.} 
for the multi-modal interactions. 
During experiments, we replace the encoder and decoder Transformer layers with the proposed LW-Transformer layer.  The model settings are similar to the standard bert~\cite{devlin2018bert}, \emph{i.e.,} $d=$768, $d_f=3156$, $d_h=96$, $k=2$, $h=6$. For simplicity, we denote the network with our optimization methods  as \textit{LW-LXMERT}.  The detailed explanation of the notations  is given in Tab.~\ref{notation}. 

During experiments, we also examine the sensitivity in mapping   dimensions of \emph{query} $\textbf{Q}$ and \emph{key} $\textbf{K}$ towards the model performance. 
Hence, we add an  suffix $n\times$ after LW-Transformer to indicate the change in attention dimensions of  $\textbf{Q}$ and  $\textbf{K}$. For instance, $d_h$ of {LW-Transformer$_{1\times}$} is 64, while the one of {LW-Transformer$_{3\times}$ } is 192.

\noindent \textbf{Swin-Transformer}: For image classification, we use Swin-Transformer-tiny~\cite{liu2021swin} as our baseline  structure. Swin-Transformer-tiny is  a hierarchical Transformer, which contains 12 Transformer blocks in 4 stages.  During experiments, we simply replace the MHA and FFN of the original Swin-Transformer-tiny   with our G-MHA and G-FFN. The modified model is denoted as LW-Transformer. To compare with the original Swin-Transformer under the same parameters,  we also build  a large model  containing 16 Transformer blocks, namely LW-Transformer-large.   Meanwhile, we remove weight-sharing in our model, since we  find that it  degrades   the performance of image classification.

\begin{table*}[t]
	\centering
	\caption{Ablation Study.  These  variations are conducted on VQA \emph{val} set, COCO Caption \emph{Karpathy} set and RefCOCO \emph{val} set, respectively.    \emph{WS} refers to \emph{weight sharing} and $k$ is the number of groups. MAdds is \emph{multiplication-addition}~\cite{howard2017mobilenets}  used to indicate the computation cost. }
	\footnotesize
	\scalebox{1.3}{\begin{tabular}{l|cccc|c|c|c|c|c}
			\toprule
			Model & G-MHA & G-FFN & WS & k & \#Params\footnotemark[3] & MAdds & VQA2.0 & COCO* & RefCOCO \\ \hline
			Transformer & - & - & - & - & 44.1M & 2.58G & 67.17 & 117.0 & 80.8 \\\hline
			LW-Transformer$_{1\times}$ & \checkmark & \xmark & \xmark & 2 & 37.0M & 2.21G & 67.11 & 116.7 & 80.9 \\
			LW-Transformer$_{1\times}$ & \xmark & \checkmark & \xmark & 2 & 37.7M & 2.22G & 67.15 & 117.1 & 81.0 \\
			LW-Transformer$_{1\times}$ & \checkmark & \checkmark & \xmark & 2 & 32.4M & 1.85G & 67.13 & 117.1 & 80.9 \\
			LW-Transformer$_{1\times}^\dagger$ & \checkmark & \checkmark & \checkmark & 2 & 24.0M & 1.85G & 67.10 & 116.9 & 80.7 \\ \hline
			LW-Transformer$_{2\times}$ & \checkmark & \checkmark & \checkmark & 2 & 26.3M & 2.16G & 67.14 & 117.1 & 80.8 \\
			LW-Transformer$_{3\times}^\dagger$ & \checkmark & \checkmark & \checkmark & 2 & 28.7M & 2.46G & 67.19 & 117.2 & 80.9 \\ \hline
			LW-Transformer$_{3\times}$ & \checkmark & \checkmark & \checkmark & 4 & 21.9M & 1.83G & 66.68 & 115.9 & 80.0 \\
			LW-Transformer$_{3\times}$ & \checkmark & \checkmark & \checkmark & 8 & 19.8M & 1.51G & 66.29 & 114.6 & 79.3 \\ \bottomrule
			\multicolumn{10}{l}{* CIDEr is used as the  metric. These results are before  the CIDEr optimization stage.}\\
			\multicolumn{10}{l}{$\dagger$ denotes the model for the further experiments.}
	\end{tabular}	}
	\label{ablation}
	\vspace{-1em}
\end{table*}

\footnotetext[3]{The parameters of the language and vision encoders, \emph{i.e.}, CNN and LSTM, are not counted.} 
\subsection{Datasets}

\noindent \textbf{Visual Question Answering}: We conduct experiments  on three vqa benchmark datasets, \emph{i.e.,} VQA2.0~\cite{goyal2017making}, GQA~\cite{hudson2019gqa} and CLEVR~\cite{johnson2017clevr}.   {VQA2.0}  contains about 1.1M image-question pairs from real word,
in which there are 440K examples for training, 210K for validation, and 450K for testing.  	{CLEVR} is a synthetically generated dataset that  aims to test the visual reasoning ability of models. It  includes 700K   and 150K  examples  for training and test, respectively.   {GQA} contains 22M questions over
140K images, which is designed to test the visual reasoning ability of models  in  real scenes.   In terms of the evaluation metric, we use the \emph{VQA accuracy}~\cite{antol2015vqa} for VQA2.0, and the classification accuracy for GQA and CLEVR.

\noindent \textbf{Referring Expression Comprehension}: 
{RefCOCO}~\cite{kazemzadeh2014referitgame} and {RefCOCO+}~\cite{kazemzadeh2014referitgame} datasets are used in our experiments. Both  datasets contain nearly 140K referring expressions for	50K bounding boxes of 20K images. The categories  of TestA are about people and the ones of TestB are about objects.  RefCOCO has more descriptions related to the  spatial relations, while  RefCOCO+ excludes these spatial-related expressions and adds more appearance-related ones.  Following \cite{hu2017modeling,hu2016natural,liu2019learning,liu2017referring,yu2018mattnet,yu2018rethinking,luo2020multi,zhou2021real}, we use  the \emph{top-1} accuracy as  the  metric on both datasets.

\noindent \textbf{Image Captioning}:  {COCO Captioning}~\cite{chen2015microsoft}   contains  more than 120K images from MS-COCO~\cite{chen2015microsoft}, each of which is annotated with 5 different captions. We train and test our models  with the \emph{Karpathy} splits \cite{karpathy2015deep}.  BLEU~\cite{papineni2002bleu}, METEOR~\cite{banerjee2005meteor}, ROUGE~\cite{rouge}, CIDEr~\cite{vedantam2015cider} and SPICE~\cite{anderson2016spice}  are used as the metrics for evaluation.

\noindent \textbf{Image Classification}:  ImageNet-1K~\cite{deng2009imagenet} is the most widely-used benchmark for image
classification, which contains
1.28M training images and 50K validation images
from 1,000 classes.  Cifar-100~\cite{krizhevsky2009learning} is a benchmark containing 60,000 low-resolution images from 100 classes. In Cifar-100, there are 50,000 images for training and 10,000 images for validation.

\begin{table*}[t]
	\centering
	\caption{Effects of deploying  Group-wise Linear-merge Layer(G-LML) and Group-wise Intermediate Layer(G-IL). The comparisons  are conducted on VQA \emph{val} set, COCO Caption \emph{Karpathy} set and RefCOCO \emph{val} set. }
	\scalebox{1.3}{	\begin{tabular}{l|cc|c|c|c|c|c}
		\toprule
		Model & G-LML & G-IL & \#Params & MAdds & VQA2.0 & COCO* & RefCOCO \\ \hline
		Transformer~\cite{vaswani2017attention} & - & - & 44.1M & 2.58G & 67.17 & 117.0 & 80.8 \\ \hline
		LW-Transformer$_{1\times}$ & \xmark & \xmark & 24.0M & 1.85G & 67.10 & 116.9 & 80.7 \\ \hline
		LW-Transformer$_{1\times}$ & \checkmark & \xmark & 20.4M & 1.69G & 66.93  & 116.5  & 80.2 \\
		LW-Transformer$_{1\times}^{mini}$ & \xmark & \checkmark & 14.5M & 1.50G & 66.57  & 115.7  & 79.8 \\
		LW-Transformer$_{1\times}$ & \checkmark & \checkmark & 11.0M & 1.33G & 66.36 & 114.8 & 79.3 \\ \bottomrule
		\multicolumn{8}{l}{* CIDEr is used as the  metric. These results are before the CIDEr optimization stage.}
	\end{tabular}}
	\vspace{-1em}
	\label{tab1}
\end{table*}

\begin{table}[t]
	\centering
	\caption{Comparisons of  LW-Transformer$_{1\times}^{mini}$ with existing multi-modal fusion networks on VQA \emph{test-dev}. All models are trained on \emph{train+val} set and tested on \emph{test-dev} set.}
	\begin{tabular}{l|c|c|c|c|c}
		\toprule
		Model & \#Params & All & Y/N & Num & \multicolumn{1}{c}{Other} \\ \hline
		MCB~\cite{fukui2016multimodal} & 32M & 61.23 & 79.73 & 39.13 & 50.45 \\
		Tucker~\cite{ben2017mutan} & \underline{14M} & 64.21 & 81.81 & 42.28 & 54.17 \\
		MLB~\cite{kim2016hadamard} & 16M & 64.88 & 81.34 & 43.75 & 53.48 \\
		MFB~\cite{yu2017multi} & 24M & 65.56 & 82.35 & 41.54 & 56.74 \\
		MUTAN~\cite{ben2017mutan} & \underline{14M} & 65.19 & 82.22 & 42.10 & 55.94 \\
		MFH~\cite{yu2018beyond} & 48M & 65.72 & 82.82 & 40.39 & 56.94 \\
		BLOCK~\cite{ben2019block} & 18M & \underline{66.41} & \underline{82.86} & \underline{44.76} & \underline{57.30} \\ \hline
		LW-Transformer$_{1\times}^{mini}$ & 14.5M & \textbf{69.68} &\textbf{86.03}  & \textbf{51.62 }& \textbf{59.80}\\ \bottomrule
	\end{tabular}
\label{mini}
	\vspace{-1em}
\end{table}

\subsection{Experiment Setups}
\noindent \textbf{Transformer}: For all datasets except CLEVR~\cite{johnson2017clevr}, we use the regional features  from \cite{anderson2018bottom} as the visual inputs. On VQA2.0, GQA and IC, all networks are trained for 13 epochs, 3 of which are for warming-up.  
The basic learning rate is set to 1e-4$\times \sqrt{k}$, where $k$ denotes the number of groups, and decayed  on 10 epochs and 12 epochs  with a factor of 0.2. For CLEVR, we follow the previous works~\cite{johnson2017clevr,hudson2018compositional,perez2018film,santoro2017simple,hu2019language} to use  grid features extracted by   ResNet101~\cite{he2016deep}.  The model is trained for up to 16 epochs and warmed up for 3 epochs. The basic learning rate is set to 4e-4$\times \sqrt{k}$ and decayed by 0.2 on 13 epochs and 15 epochs. For IC,  another 17 training epochs  is  further used for  the CIDEr-D~\cite{vedantam2015cider,liu2017improved} optimization with a learning rate of 5e-6. The Adam optimizer~\cite{kingma2014adam} is used  to train all networks.

\noindent \textbf{Bert-style Model}: Following the settings of LXMERT~\cite{tan2019lxmert}, the visual features are extracted by Faster R-CNN~\cite{ren2015faster} with ResNet101~\cite{he2016deep} backbone  pretrained on the Visual Genome~\cite{krishna2017visual}. The training procedures are divide into two steps, \emph{i.e.,} pretraining and finetune.  We follow the default setting of LXMERT to pre-train the model. The pretraining takes  20 epochs overall, where the optimizer is  Adam ~\cite{kingma2014adam} with a initial learning rate of 1e-4$\times \sqrt{k}$.  We then finetune the model on VQA and GQA  with a learning rate of 1e-5$\times \sqrt{k}$ for 5 epochs.

\noindent \textbf{Swin-Transformer}: Following the default settings of Swin-Transformer~\cite{liu2021swin}, we use the AdamW optimizer with an initial learning rate of
0.001 and a weight decay of 0.05.
We use a cosine decay learning rate scheduler with 20 epochs of linear
warm-up.   Models are trained for total 300 epochs with 1024 batchsize.   For ImageNet and Cifar-100,  we adopt the input image resolution of $224 \times 224$. Following the training strategy of Swin-Transformer, a set of data augmentations, \emph{i.e.,} RandAugment, Mixup, Cutmix, random erasing and stochastic depth, are applied to avoid overfitting.

\begin{table*}[t]
	\centering
	\caption{Comparisons of LW-Transformer between Group-wise Transformation (LW-Transformer) and efficient transformers~\cite{ma2019tensorized,choromanski2020rethinking,wang2020linformer}. The comparisons are conducted on VQA \emph{val} set, COCO Caption \emph{Karpathy} set and RefCOCO \emph{val} set, respectively. MAdds \emph{i.e.,} multiplication-additions~\cite{howard2017mobilenets}, is used to indicate the computation cost.  }
	
	\scalebox{1.3}{\begin{tabular}{l|c|c|c|c|c}
			\toprule
			Model & \#Params\footnotemark[3]& \multicolumn{1}{l|}{MAdds} & VQA2.0 & \multicolumn{1}{l|}{COCO*} & \multicolumn{1}{l}{RefCOCO} \\ \hline
			Transformer~\cite{vaswani2017attention} & 44.1M & 2.58G & 67.17 & 117.0 &80.8  \\
			Tensorized Transformer~\cite{ma2019tensorized} & 27.5M &  3.71G & 67.03 &116.3  &   80.6\\
			Performer~\cite{choromanski2020rethinking}&44.1M&3.36G&65.20 &	116.2&	80.0
			\\
			Linformer~\cite{wang2020linformer}&44.2M&2.53G&64.20&	-	&80.3
			\\
			LW-Transformer$_{1\times}$ & \textbf{24.0M }& \textbf{1.85G} & 67.10 & 116.9 &  80.7 \\
			LW-Transformer$_{3\times}$ & 28.7M & 2.46G & \textbf{67.19 }& \textbf{117.2} &  \textbf{80.9}\\ \bottomrule
			\multicolumn{6}{l}{* CIDEr is used as the  metric. These results are before the CIDEr  }\\
			\multicolumn{6}{l}{optimization stage.}
	\end{tabular}}
		\vspace{-1em}
	\label{comp}
\end{table*}

\begin{table*}[t]
	\centering
	\caption{Comparisons of LW-Transformer and SOTAs on VQA tasks in single-model setting.}
	\scalebox{1.3}{	\begin{tabular}{lc|l|c|l|c}
			\toprule
			\multicolumn{2}{c|}{VQA2.0} & \multicolumn{2}{c|}{CLEVR} & \multicolumn{2}{c}{GQA}  \\ \cline{1-6}
			\multicolumn{1}{l|}{model} & test-dev & model & test & model & test-dev   \\ \hline
			\multicolumn{1}{l|}{Bottom-Up~\cite{anderson2018bottom}} & 65.3 & SAN~\cite{johnson2017clevr} & 76.7 & Bottom-Up~\cite{anderson2018bottom} & 49.7  \\
			\multicolumn{1}{l|}{MFH~\cite{yu2018beyond}} & 68.8 & RN~\cite{santoro2017simple} & 95.5 & MAC~\cite{yu2019deep} & 54.1  \\
			\multicolumn{1}{l|}{BAN~\cite{kim2018bilinear}} & 70.0 & FiLM~\cite{perez2018film} & 97.7 & LCGN~\cite{hu2019language} & \underline{57.1}  \\
			\multicolumn{1}{l|}{MCAN~\cite{yu2019deep}} & \underline{70.6} & MAC~\cite{hudson2018compositional} & \underline{98.9} & BAN~\cite{kim2018bilinear} & \underline{57.1} \\ \hline
			\multicolumn{1}{l|}{Transformer~\cite{vaswani2017attention}} & \textbf{70.6} & Transformer~\cite{vaswani2017attention} & 98.4 & Transformer~\cite{vaswani2017attention} & 57.4  \\\hline
			\multicolumn{1}{l|}{LW-Transformer$_{1\times}$ } & 70.4 & LW-Transformer$_{1\times}$  & 98.6 & LW-Transformer$_{1\times}$  & \textbf{58.4}  \\
			\multicolumn{1}{l|}{LW-Transformer$_{3\times}$ } & 70.5 & LW-Transformer$_{3\times}$  & \textbf{98.7} & LW-Transformer$_{3\times}$  & 57.5 \\ \bottomrule
	\end{tabular}}
	\label{tab4}
		\vspace{-1em}
\end{table*}

\begin{table*}[t]
	\centering
	\caption{Comparisons of LW-Transformer and SOTAs on REC tasks in single-model setting.}
	\scalebox{1.3}{	\begin{tabular}{l|c|c|lcc}
			\toprule
			\multicolumn{3}{c|}{RefCOCO} & \multicolumn{3}{c}{RefCOCO+}  \\ \cline{1-6}
			\multicolumn{1}{c|}{model} & testA & testB & \multicolumn{1}{c|}{model} & \multicolumn{1}{c|}{testA} & testB  \\ \hline
			Spe+\textbf{Lis}+Rl~\cite{yu2017joint} & 73.1 & 64.9 & \multicolumn{1}{l|}{Spe+\textbf{Lis}+Rl~\cite{yu2017joint}} & \multicolumn{1}{c|}{60.0} & 49.6  \\
			DDPN~\cite{yu2018rethinking} & 80.1 & 72.4 & \multicolumn{1}{l|}{DDPN~\cite{yu2018rethinking}} & \multicolumn{1}{c|}{70.5} & 54.1  \\
			MattNet~\cite{yu2018mattnet} & 81.1 & 70.0 & \multicolumn{1}{l|}{MattNet~\cite{yu2018mattnet}} & \multicolumn{1}{c|}{71.6} & 56.2 \\
			NMTree~\cite{liu2019learning} & \underline{81.2} & \underline{70.1} & \multicolumn{1}{l|}{NMTree~\cite{liu2019learning}} & \multicolumn{1}{c|}{\underline{72.0}} & \underline{57.5}  \\ \hline
			Transformer~\cite{vaswani2017attention} & 84.0 & 73.4 & \multicolumn{1}{l|}{Transformer~\cite{vaswani2017attention}} & \multicolumn{1}{c|}{75.9} & \textbf{61.1} \\ \hline
			LW-Transformer$_{1\times}$  & \textbf{84.2} & 73.7 & \multicolumn{1}{l|}{LW-Transformer$_{1\times}$ } & \multicolumn{1}{c|}{75.9} & 61.0 \\
			LW-Transformer$_{3\times}$  & 83.9 & \textbf{74.3} & \multicolumn{1}{l|}{LW-Transformer$_{3\times}$ } & \multicolumn{1}{c|}{\textbf{76.5}} & 60.8  \\\bottomrule
	\end{tabular}}
	\label{tab5}
		\vspace{-1em}
\end{table*}

\begin{table*}[t]
	\centering
	\caption{Comparisons of LW-Transformer and SOTAs on Image Captioning tasks in single-model setting. The models are evaluated  on the \emph{Karpathy} test split.}
	\scalebox{1.3}{	\begin{tabular}{l|c|c|c|c|c|c}
			\toprule
			\multicolumn{6}{c}{COCO Captioning} \\ \hline
			\multicolumn{1}{c|}{model} & Params & BLEU-4 & METEOR & ROUGE & CIDEr & SPICE \\ \hline

			ORT~\cite{herdade2019image}&45M & 38.6 & 28.7 & 58.4 & 128.3 & {22.6} \\
			AoANet~\cite{huang2019attention}&64M & 38.9 & {29.2} & {58.8} & 129.8 & 22.4 \\
						GCN-LSTM+HIP~\cite{yao2019hierarchy} &- &39.1 &28.9& \underline{59.2} &130.6 &22.3\\
			$\mathcal{M}^2$ Transformer~\cite{cornia2019m}&33M & {39.1} & {29.2} & 58.6 & {131.2} & {22.6} \\
			X-Transformer~\cite{pan2020x}&138M& \underline{39.7}& \underline{29.5}& {59.1}& \underline{132.8}& \underline{23.4}\\ \hline
			Transformer~\cite{vaswani2017attention}&44M & {38.9} & 29.0 & 58.5 & 131.0 & 22.3 \\\hline
			LW-Transformer$_{1\times}$ &24M & 38.7 & \textbf{29.2} & 58.3 & 130.9 & \textbf{22.7} \\
			LW-Transformer$_{3\times}$ &29M  & \textbf{38.9} & \textbf{29.2} & \textbf{58.6} & \textbf{131.3} & 22.6\\\bottomrule
	\end{tabular} }
	\label{tab6}
	\vspace{-2mm}
\end{table*}

\subsection{Experiment Results}

\subsubsection{Ablation Study} We first examine different designs  of the proposed LW-Transformer  including \emph{group-wise multi-head attention} (G-MHA), \emph{group-wise feed-forward network} (G-FFN) and \emph{weight sharing} (WS), and also evaluate the sensitivity of hyper-parameters like the group number ($k$) and the dimensions of \emph{query} and \emph{key} in MHA (indicated by the subfix $n\times$ after LW-Transformer, \emph{i.e.}, $d_h$  defined in Eq.~\ref{mha_func}).  The results are reported in Tab.~\ref{ablation}.  

\textbf{The effect of group-wise transformation.} The second block of Tab.~\ref{ablation} illustrates the impact of G-MHA, G-FFN and WS on the parameter size, computation cost and overall performance  of LW-Transformer. 
From these results we can see that while  greatly reducing the   model parameters  and computations, each of these designs will hardly reduce model performance, and even help the model obtain slightly improvements on COCO and RefCOCO. 
In addition, after  deploying all  designs, \emph{i.e.,}  LW-Transformer$_{1\times}$,  the parameters and computations  are reduced by 45.5\% and 28.3\%, respectively. However, the performance drop is still very marginal, {\textit{e.g.,} -0.1\% on VQA2.0, -0.08\% on COCO and -0.12\% on RefCOCO.  }  Thus, we can conclude that  group-wise designs  in LW-Transformer can effectively reduce both  parameter size and computation cost, while  maintaining the  performance.

\textbf{The impact of expanding attention dimensions.} We  explore the effects of expanding the attention dimensions of \emph{query} and \emph{key} in MHA, of which results are given in the third block of Tab.\ref{ablation}. 
Here, the suffixes, \emph{i.e.,} $\textbf{1}\times$ and $\textbf{3}\times$, indicate the multipliers of dimension expansion. 
From these results, we  observe that expanding the MHA dimension  further improve the model performance, which helps LW-Transformer outperform  the original Transformer on all datasets    with much fewer parameters.  For instance, compared with LW-Transformer$_{1\times}$, the efficiency of LW-Transformer$_{2\times}$ is  relatively less significant, but it still reduce 40\% parameters and 16.3\% computations, respectively. More importantly, it has almost the same performance to the default Transformer. 
We also notice that the increased parameter size is still small, \emph{e.g.}, +4.7M for $3\times$. 
 In contrast,   expanding the projection dimensions  significantly increases the model size  in the original MHA, \emph{e.g.}, +18.8M for $3\times$. 
To explain,  the input features in G-MHA are first truncated into multiple groups, which will leads to less parameters than the direct projection. 
 This  observation  also  confirms another merit of  Group-wise transformation.

\textbf{The impact of group number.} Next, we test LW-Transformer with different group numbers ($k$), as reported in the last block of Tab.~\ref{ablation} . 
From this table, we notice that although increasing the group number  further reduces the model size, the performance degradation  becomes relatively   obvious. 
For this case, our understanding is that both the attention modeling and feature transformation  require a certain capacity  to learn the large-scale language-and-vision data, \emph{e.g.}, VQA2.0,  and  setting  too small  feature  dimensions will be counterproductive.
In addition, since some layers  are not included in our  optimization  scheme, \emph{e.g.}, the linear merge layer in MHA,  the lower bound of LW-Transformer's parameter size is about 20M. 	After experiments, we find that a  trade-off between the efficiency and the performance is $k=2$. 

\begin{table*}[t]
	\centering
	\caption{Comparisons between LW-Transformer and other Transformers on image classification task. We use Swin-Transformer-tiny~\cite{liu2021swin} as the baseline architecture. LW-Transformer-large denotes the model with more Transformer layers.  }
	\begin{tabular}{l|ccc|cc|cc}
		\toprule
		\multicolumn{1}{c|}{\multirow{2}{*}{\textbf{\begin{tabular}[c]{@{}c@{}}Model\end{tabular}}}} & \multirow{2}{*}{\textbf{groups}} & \multirow{2}{*}{\textbf{params}} & \multirow{2}{*}{\textbf{MAdds}} & \multicolumn{2}{c|}{\textbf{Cifar-100}} & \multicolumn{2}{c}{\textbf{ImageNet}} \\ \cline{5-8} 
		\multicolumn{1}{c|}{}                                                                                         &                                  &                                  &                                 & top-1              & top-5              & top-1              & top-5             \\ \hline
		Swin-Transformer~\cite{liu2021swin}                                                                                                   & -         & 28.3M     & 4.5G      & 77.9               & 94.5               & 81.2               & 95.5              \\
		Performer~\cite{choromanski2020rethinking}                                                                                                     & -         & 27.5M     & 4.1G      & -                  & -                  & 79.0               & 94.2                \\ 
		LW-Transformer$_{1\times}$                                                                                                 & 2         & 20.0M     & 3.3G      & 77.8               & 94.7               & 79.9               & 94.9              \\
		LW-Transformer-large$_{1\times}$& 2&27.0M&4.5G&78.8&95.2&81.5&95.7 \\
		LW-Transformer$_{1\times}$                                                                                                 & 4         & 16.9M     & 2.6G      & 77.7               & 94.5               & 78.9               & 94.4              \\ \bottomrule
	\end{tabular}
\label{classification}
\vspace{-2mm}
\end{table*}

\begin{table*}[t]
	\centering
	\caption{A statistic summary of the effects of our solutions on 5 Transformer networks.}
	\footnotesize
	\setlength\tabcolsep{3.0pt}
\begin{tabular}{l|ccc|cc|c|c|c}
	\toprule
	\multicolumn{1}{c|}{\multirow{3}{*}{Metrics}} & \multicolumn{3}{c|}{\multirow{2}{*}{\begin{tabular}[c]{@{}c@{}}Visual Question\\ Answering\end{tabular}}} & \multicolumn{2}{c|}{\multirow{2}{*}{\begin{tabular}[c]{@{}c@{}}Reffering Expression\\ Comprehension\end{tabular}}} & \multirow{2}{*}{\begin{tabular}[c]{@{}c@{}}Image\\ Captioning\end{tabular}} & \multirow{2}{*}{\begin{tabular}[c]{@{}c@{}}V\&L\\ Pre-training\end{tabular}} & \multirow{2}{*}{\begin{tabular}[c]{@{}c@{}}Image\\ Classification\end{tabular}} \\
	\multicolumn{1}{c|}{}                         & \multicolumn{3}{c|}{}                                                                                     & \multicolumn{2}{c|}{}                                                                                              &                                                                             &                                                                              &                                                                                 \\ \cline{2-9} 
	\multicolumn{1}{c|}{}                         & \multicolumn{1}{c|}{VQA2}                & \multicolumn{1}{c|}{CLEVR}               & GQA                 & \multicolumn{1}{c|}{RefCOCO}                                       & RefCOCO+                                      & MS-COCO                                                                     & GQA                                                                          & Cifar100                                                                        \\ \hline
	Performance Gains                             & \multicolumn{1}{c|}{-0.1}                & \multicolumn{1}{c|}{+0.3}                & +1.0                & \multicolumn{1}{c|}{+0.2}                                          & +0.0                                          & -0.1                                                                        & +0.6                                                                         & -0.1                                                                            \\
	Parameter Reduction                           & \multicolumn{1}{c|}{-45\%}               & \multicolumn{1}{c|}{-45\%}               & -45\%               & \multicolumn{1}{c|}{-45\%}                                         & -45\%                                         & -45\%                                                                       & -45\%                                                                        & -41\%                                                                           \\
	Computational Reduction                       & \multicolumn{1}{c|}{-28.3\%}             & \multicolumn{1}{c|}{-28.3\%}             & -28.3\%             & \multicolumn{1}{c|}{-28.3\%}                                       & -28.3\%                                       & -28.3\%                                                                     & -28.3\%                                                                      & -42.2\%                                                                         \\ \bottomrule
\end{tabular}
\end{table*}
\begin{table}[t]
	\centering
	\caption{Comparisons  between Bert-style models on the \emph{test-dev} splits of VQA2.0 and GQA. LW-LXMEERT is the model deployed with the proposed LW-Transformer layer. }
	\scalebox{1.2}{	\begin{tabular}{l|c|c|c}
			\toprule
			model & VQA2.0 & GQA & \#Params\footnotemark[3] \\ \hline
			ViLBERT~\cite{lu2019vilbert} & 70.6 &-& 221M\\
			VisualBERT~\cite{li2019visualbert} & 70.8 & - & \underline{85M} \\ 
			UNITER~\cite{li2019unicoder} & 72.3 & - & \underline{85M} \\
			12in1~\cite{lu201912} & \underline{72.6} & \underline{60.1} & 221M \\ \hline
			LXMERT~\cite{tan2019lxmert} & \textbf{72.5} & 60.3 & 181M \\\hline
			LW-LXMERT$_{1\times}$ (Ours)  & 71.6 & \textbf{60.6} & \textbf{100M} \\
			LW-LXMERT$_{3\times}$ (Ours) & 71.8 & 60.5 & 117M \\ \bottomrule
	\end{tabular}}
	\label{tab7}
	\vspace{-2mm}
\end{table}
\subsubsection{Fully Group-wise LW-Transformer}
In  LW-Transformer,  Group-wise transformation is not applied in two inner layers of Transformer, \emph{i.e.,} \emph{linear merge layer}  of MHA and the \emph{ intermediate layer} of FFN. In practice, we find that these   two layers are important for keeping the model capacity and facilitating the cross-group information exchange. In Tab.~\ref{tab1}, we quantitatively validate the effects of deploying Group-wise transformation on these two layers, denoted as Group-wise Linear-Merge Layer (G-LML) and Group-wise Intermediate Layer (GIL), respectively. From Tab.~\ref{tab1}, we can see that when  G-LML and G-IL are deployed, the parameters and computations are further reduced by 29.5\% and 19.7\%, respectively, but the performance declines are relatively  significant, \emph{e.g.,} from 67.10 to 66.36 on VQA and from 116.9 to 114.8 on COCO. To explain,  the completely group-wise structure  divides the network into two independent sub-networks without any cross-group information exchange, which ultimately results in the performance degeneration.  Based on these observations, we therefore keep the linear-merge layer of MHA and the intermediate layer of FFN intact  in LW-Transformer.

In addition, we also notice that retaining G-IL is the best choice after the design introduced in the paper, where both compactness and performance are satisfactory. Therefore, we further extend LW-Transformer to a mini version, \emph{i.e.},  deploying G-IL and keeping the linear-merge layer intact  for cross-group information exchange, termed as LW-Transformer$^{mini}$  and  compare it with the   existing multi-modal fusion networks~\cite{ben2017mutan,ben2019block,fukui2016multimodal,kim2016hadamard,yu2017multi,yu2018beyond}, of which results are shown in Tab.~\ref{mini}. From Tab.~\ref{mini}, we find that even compared to  state-of-the-art methods, \emph{i.e.,} BLOCK~\cite{ben2019block},  LW-Transformer$_{1\times}^{mini}$  outperforms it by a large margin, while  the number of parameters is  much smaller, \emph{i.e.}, 14.5M \emph{vs} 18M. Such results  greatly support the effectiveness of LW-Transformer  under the extremely lightweight settings.   Meanwhile, it also suggests that the liner merge layer (LML) is an important design in MHA, which needs to be carefully considered during optimization.

\subsubsection{Comparisons with  other Efficient  Transformers}
In Tab.~\ref{comp}, we also compare  LW-Transformer with existing efficient transformers including Tensorized Transformer~\cite{ma2019tensorized}, Performer~\cite{choromanski2020rethinking} and Linformer~\cite{wang2020linformer}. 
In Tab.~\ref{comp}, Tensorized Transformer applies \emph{Tucker Decomposition}, a popular method in network compression~\cite{ma2019tensorized}, to decompose the projection  weights in MHA. {Performer~\cite{choromanski2020rethinking} approximates the scaled-dot product attention by using scalable kernel methods. Linformer~\cite{wang2020linformer} uses additional projection layers to reduce the length of the input sequence, which is hard to apply to apply to the sequence-to-sequence generation tasks like Image Captioning. So, we only report  the results  of Linformer  on VQA2.0 and RefCOCO. }

From  Tab.~\ref{comp},  we can see that  LW-Transformer obtains better performance than the other compressing methods, and its performance gains are very distinct on VQA2.0. 
{For these results, our understanding is that   changing the original definition of self-attention will affect the effectiveness of Transformer to some degree. For instance, Linformer reduces the length of the features for attention, which inevitably leads to the loss of information for fine-grained tasks like VQA. The performance of Performer also suggests that there is still a gap between the approximated attention and the default one.  }

{Secondly, compared with the original structure, Performer and Tensorized Transformer  greatly increases the computation cost  (+30\% and +43.8\%), while our methods can reduce it by up to 28.3\%.   The merit of Linformer in computation is also very limited. 
	To explain,	  Linformer and Performer are  proposed for the tasks with very large input sequences,\textit{ e.g., }language modeling~\cite{chelba2013one} and unidirectional/causal modeling~\cite{parmar2018image}. In unidirectional/causal modeling, the sequence length of input features can be up to 12,288, so the computation reductions by Linformer and Performer can be up to 68.8\% and 41.3\%, respectively. In contrast, the sequence length for VL tasks typically ranges from 30 to 100~\cite{ goyal2017making,kazemzadeh2014referitgame,chen2015microsoft}, which are much smaller.  Therefore, these  efficient transformer will increase the amount of computations  for VL tasks in contrast.   }

From these observations, we can conclude that the efficiency of LW-Transformer is indeed obvious, especially considering its performance gains to other  optimization  methods. More importantly, it can improve the efficiency of Transformer while maintaining the  original definition of attention modeling.

\subsubsection{Comparisons with the State-of-the-art Methods}
We further compare LW-Transformer with SOTA methods on six benchmark datasets of three multi-modal tasks, \emph{i.e.}, VQA, IC and REC. The  results are given in Tab.~\ref{tab4} - \ref{tab6}.

Tab.~\ref{tab4} shows the comparison on three widely-used benchmark datasets of VQA, \emph{i.e.}, VQA2.0~\cite{goyal2017making}, CLEVR~\cite{johnson2017clevr} and GQA~\cite{hudson2019gqa}. 
The first observation from this table is that the performance of the proposed LW-Transformer is very competitive on all three datasets. 
And it   achieves  new SOTA performance on GQA, which is the largest VQA dataset for  visual reasoning. 
Besides, compared to these SOTA methods, LW-Transformer has an obvious advantage in terms of the parameter size. For instance, BAN-8 and MCAN have 93.4 and 44.1 millions of parameters\footnotemark[3], respectively, while LW-Transformer$_{1\times}$ and LW-Transformer$_{3\times}$ only have 24.0 and 28.7 millions of parameters, respectively.  
In addition, we  observe that LW-Transformer achieves better performance than Transformer except for VQA2.0, which has a strong language bias and requires a larger parameter size to accommodate the data distribution~\cite{goyal2017making}.
These observations greatly confirm the effectiveness of LW-Transformer and the introduced Group-wise Transformation on the VQA task. 

On REC, which is a task of exploring  language-vision alignment, the advantages of LW-Transformer   are more significant, as shown in Tab.~\ref{tab5} . 
Compare to SOTA methods like NMTree~\cite{liu2019learning}, the performance gains of LW-Transformer  are up to 6.1\% (on RefCOCO+), greatly showing the generalization ability of LW-Transformer. 
We  also observe that its performance is  very close to that of Transformer, and even slightly better on RefCOCO, which  once again confirms our argument  that Group-wise transformation can compress Transformer while still keeping its performance.

\begin{figure*}[t]
	\centering
	\includegraphics[width=2\columnwidth]{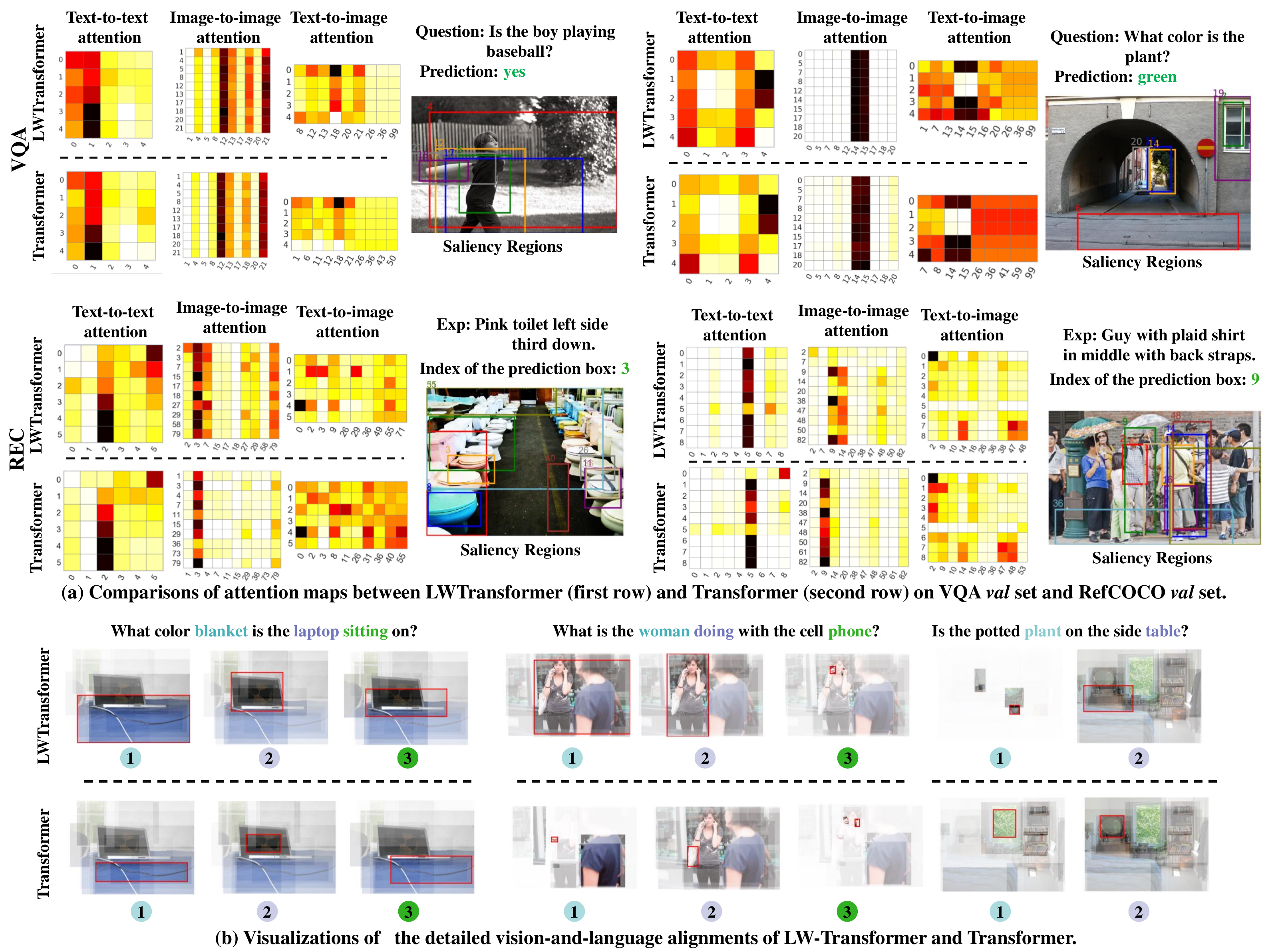}
	
	\caption{  Visualizations of the attention maps of the proposed  LW-Transformer and  the default  Transformer. We visualize  different types of attention maps  in (a) and  the detailed vision-and-language alignments  in (b).  These results are all from the last self-attention layer of two models. In (b), the red box denotes the most attended visual region.  }
	\label{vis1}
	\vspace{-1em}
\end{figure*}
The comparison on the language-and-vision generative task, \emph{i.e.}, IC, is given in Tab.~\ref{tab6} . 
As  shown in  this table,  LW-Transformer  is  an excellent generative model. Its performance is very close to 	$\mathcal{M}^2$  Transformer, and even better on the metrics of CIDEr and SPICE. 
In addition, the  parameter size of LW-Transformer is     smaller than that of 	$\mathcal{M}^2$ Transformer, \emph{i.e.}, 24 millions  \emph{vs} 33 millions. Considering M2 Transformer is already an efficient model, these improvements greatly suggest the effectiveness and efficiency of LW-Transformer.    {When compared to the SOTA method, \emph{i.e.,} X-Transformer, the advantage of LW-Transformer in efficiency becomes more prominent. Compared with X-Transformer, LW-Transformer has up to 79.0\% fewer parameters, while the performance is only reduced by 1.12\% on CIDEr.  }

\subsubsection{Quantitative Results with the BERT-style Pre-training}
We further apply   our optimization strategy  to a BERT-style model, \emph{i.e.}, LXMERT~\cite{tan2019lxmert}, and compare it with a set of methods  pre-trained on the large-scale language-and-vision data. The result  is given in Tab.~\ref{tab7}.
For simplicity, we denote the compressed LXMERT as LW-LXMERT.  
From this table, we  first observe that  after  deploying  G-MHA and G-FFN, the parameter size of LXMERT   is reduced by up to 44.8\%, while the performance is marginally reduced on VQA2.0 and even better on GQA. 
We  notice that compared to some SOTA methods like 12in1~\cite{lu201912}, the performance of LW-LXMERT on VQA2.0 is  slightly worse. 
As mentioned about, VQA2.0 requires a larger model capacity due to  the issue of  language bias~\cite{goyal2017making}. 
On GQA,  examples of  which are more balanced and more challenging, LW-LXMERT achieves  new SOTA performance. 
Such results confirm again our argument that LW-Transformer can be applied to most Transformer-based networks,  while maintaining their high performance.

\subsubsection{Results of Image Classification}
{ To further examine the generalization ability of LW-Transformer, we build it on  the recently proposed Swin-Transformer~\cite{liu2021swin}  and   conduct additional experiments on the task of image classification, of which results are given in Tab.~\ref{classification}. } In this table, we report three LW-Transformer with different settings. The first two are LW-Transformer$_{1\times}$ with different group numbers, and the last one is LW-Transformer$_{1\times}$-large, which has similar parameter size as Swin-Transformer by adding more Transformer layers. 

{From these results, we  have some observations. Firstly,  on Cifar-100, LW-Transformer$_{1\times}$ hardly degrades the performance of Swin-Transformer. On \textit{top-5 accuracy}, LW-Transformer$_{1\times}$  even performs slightly better. These results are consistent with those in vision-and-language benchmarks.  Secondly, on  the large-scale ImageNet benchmark, LW-Transformer$_{1\times}$ is slightly worse than Swin-Transformer. In this regard, our assumption is that ImageNet has a higher requirement for model capacity. Considering the saved experimental expenditure, e.g., saving 40.3\% parameters and 42.2\% MAdds when group number is 4, the performance drop (-2.8\%) is still acceptable.  This assumption is validated in the results of LW-Transformer-large.  It  improves Swin-Transformer on Cifar-100 and ImageNet with similar parameters and computations. Considering Swin-Transformer is carefully designed for image classification, these slightly improvements well validate the effectiveness of our method.
	Lastly, compared with Performer, LW-Transformer$_{1\times}$ has all-round advantages. Retaining better performance, LW-Transformer$_{1\times}$  merits in the parameter size and the computation cost, which greatly validates our motivation about the application of group-wise transformation.  }  {Overall, these results well support the generalization ability and efficiency  of LW-Transformer  for the traditional computer vision tasks. }


\subsection{Qualitative Experiment.} {To gain deep insights  into LW-Transformer, we   visualize its attention maps and compare  them  with  the ones of  the default  Transformer in Fig.\ref{vis1}.}
	
In Fig.~\ref{vis1}~(a), we present the overall patterns of single- and multi-modal attentions  learned by    Transformer and the proposed LW-Transformer.  From these visualizations, we can see that the global dependency patterns learned by two networks are roughly the same, which subsequently confirms the argument we made in this paper, \emph{i.e.,} group-wise transformation essentially inherits the principle of self-attention.
	
In addition to this  observation, we   capture some subtle differences  between  two networks and two multi-modal tasks. For example, the attention patterns of VQA and REC are slightly different. As shown in Fig.~\ref{vis1}~(a), the words indicating question type will be more important in VQA, \emph{e.g.}, ``\textit{is the}'' and ``\textit{what color}''.  In contrast, REC  requires the model to focus more on spatial  information  like `` \textit{left}’’ or `` \textit{middle}’’.    This result further reflects the difference  of model reasoning between two tasks. That is, VQA relies more on the language prior of question types to answer the question, and REC locates the target instance based more on spatial information. Nevertheless, we  find that these different attention patterns can be mastered  by both  LW-Transformer and Transformer. 

	In terms of vision-and-language alignment, the attention focus of LW-Transformer is more concentrated, while the one of Transformer is more divergent. For instance, in the second example of VQA, Transformer’s attention is flat, mainly focusing on the incorrect text-image relationships (4-7). In contrast, the attention of LW-Transformer is more accurate.  These properties is also reflected in the visualizations of cross-modal attentions in Fig.\ref{vis1}.b.
{
 From the examples of Fig.~\ref{vis1}.b, we can see that LW-Transformer can obtain more precise ``attention’’ than Transformer. Specifically, in cross-modal attention, Transformer and LW-Transformer focus on similar image regions. However, LW-Transformer will be more accurate in terms of the most attended regions (with red boxes), for example, the ``phone’’ in the second example. This observation confirms the merits of LW-Transformer in vision-and-language alignment.  }

\section{Conclusion}
	In this paper, we introduce the \emph{Group-wise Transformation} to achieve a lightweight yet general Transformer  network, termed \emph{LW-Transformer}. 
Compressing Transformer still retains challenging mainly due to its complex layer designs, \emph{i.e.}, MHA and FFN. 
The  proposed LW-Transformer  can well maintain their main principles, \emph{i.e.}, the efficient attention modelings on diverse subspaces and the expanding and scaling feature transformation, and reduce the parameter sizes and computation costs to a large extend.  More importantly, the intention of group-wise transformation is consistent  with MHA, which can ensure LW-Transformer to learn similar or even better attention patterns as the default Transformer. 
To validate  our optimization strategy, we  build LW-Transformer  based on Transformer and a BERT-style model called LXMERT,  and  conduct extensive experiments on six benchmark datasets of three multi-modal tasks.
The experimental results show that while saving up to 45\% of parameters and 28\% of computation costs, LW-Transformer can achieve almost or even better performance than Transformer  on these datasets.  Meanwhile, the generalization ability of LW-Transformer is also well validated on a newly proposed image Transformer called Swin-Transformer~\cite{liu2021swin} for the task of image classification.  These results greatly validate the effectiveness of LW-Transformer as well as  our motivation.


%

\appendices
%

\section*{Acknowledgment}
This work is supported by the National Science Fund for Distinguished Young Scholars (No.62025603), the National Natural Science Foundation of China (No.U1705262, No. 62072386, No. 62072387, No. 62072389, No.62002305,  
No.61772443, No.61802324 and No.61702136), China Postdoctoral Science Foundation (2021T40397), Guangdong Basic and Applied Basic Research Foundation (No.2019B1515120049) and the Fundamental Research Funds for the central universities (No. 20720200077, No. 20720200090 and No. 20720200091). 


\ifCLASSOPTIONcaptionsoff
  \newpage
\fi



\bibliographystyle{IEEEtran}
\bibliography{IEEEabrv,mybstfile}
%

%

\begin{IEEEbiography}[{\includegraphics[width=1in,height=1.25in,clip,keepaspectratio]{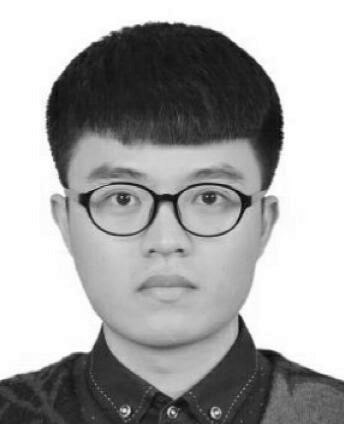}}]{Gen Luo}  is currently pursuing the phd’s
	degree in Xiamen University.
	His research interests include vision-and-language interactions.
\end{IEEEbiography}

\begin{IEEEbiography}[{\includegraphics[width=1in,height=1.25in,clip,keepaspectratio]{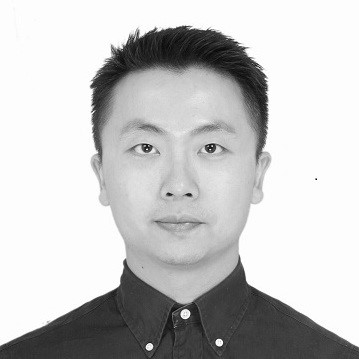}}]{Yiyi Zhou}
   received his Ph.D. degree supervised by Prof. Rongrong Ji from Xiamen University, Chian, in 2019. He is a Post-doctoral Research Fellow of the School of Informatics and a member of Media Analytics and Computing (MAC) lab of Xiamen University, China. 
\end{IEEEbiography}

\begin{IEEEbiography}[{\includegraphics[width=1in,height=1.25in,clip,keepaspectratio]{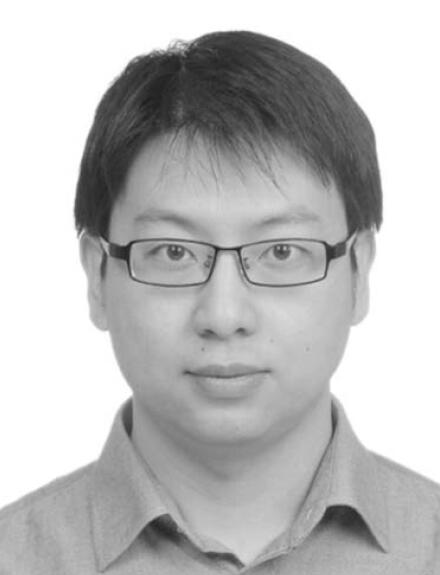}}]{Xiaoshuai Sun} (Senior Member, IEEE)
received the B.S. degree in computer
science from Harbin Engineering University, Harbin,
China, in 2007, and the M.S. and Ph.D. degrees in
computer science and technology from the Harbin
Institute of Technology, Harbin, in 2009 and 2015,
respectively. He was a Postdoctoral Research Fellow
with the University of Queensland from 2015 to
2016. He served as a Lecturer with the Harbin
Institute of Technology from 2016 to 2018. He is
currently an Associate Professor with Xiamen University,
China. He was a recipient of the Microsoft Research Asia Fellowship in 2011.
\end{IEEEbiography}

\begin{IEEEbiography}[{\includegraphics[width=1in,height=1.25in,clip,keepaspectratio]{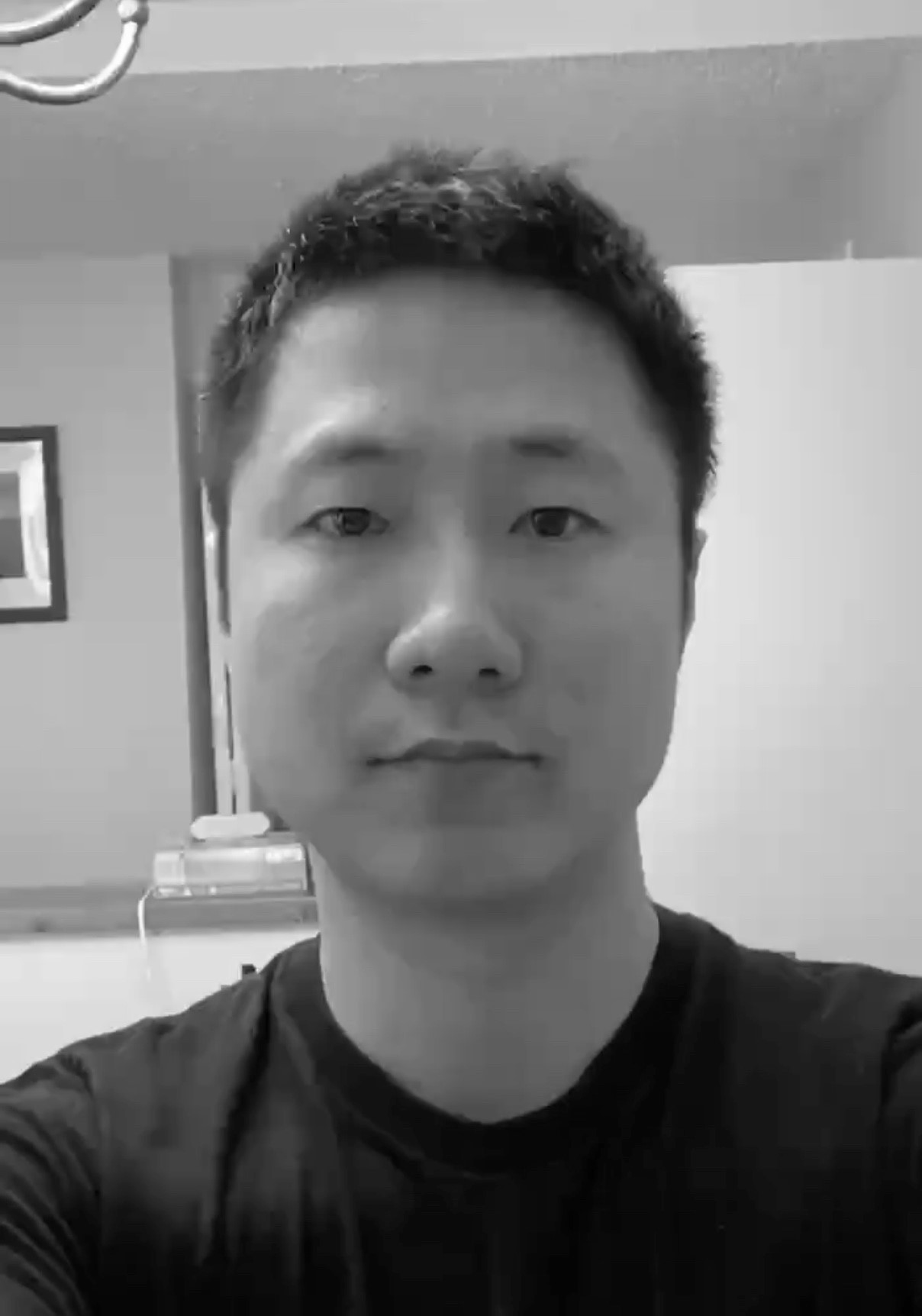}}]{Yan Wang}
got his PhD degree in Oct. 2015, under supervision of Prof. Shih-Fu Chang. He had rich R\&D experience in Adobe Research, Microsoft Research, and Facebook, with his algorithms integrated in Facebook Graph Search and Adobe Photoshop, and granted patents. He held the Olympic Torch of Beijing Olympic Games as a torchbearer in 2008, won the Tech Draft (a nation-wide programming challenge) in 2014, and is a certified airplane pilot. His current research interests include computer vision and machine learning.

\end{IEEEbiography}

\begin{IEEEbiography}[{\includegraphics[width=1in,height=1.25in,clip,keepaspectratio]{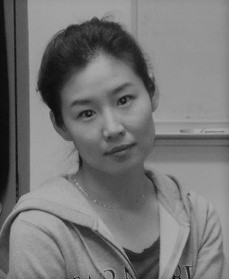}}]{Liujuan Cao} received her Bachelor’s, Master’s, and Ph.D. degrees from the School of Computer Science and Technology, Harbin Engineering University, Harbin, China. She was a Visiting Researcher with Columbia University from 2012 to 2013. She joined Xiamen University in 2014. She is currently an Associate Professor with the School of Informatics, Xiamen University, Xiamen, China. She has published more than 30 papers in the top and major tiered journals and conferences, including IEEE CVPR, Information Sciences, Neurocomputing, Signal Processing, Digital Signal Processing, etc. Her research interests include covers pattern recognition, data mining, and computer vision. She is the Finance Chair of IEEE MMSP 2015. She has been project PI for various projects including NSFC, military projects, with over 1M RMB fundings.
	
\end{IEEEbiography}
\begin{IEEEbiography}[{\includegraphics[width=1in,height=1.25in,clip,keepaspectratio]{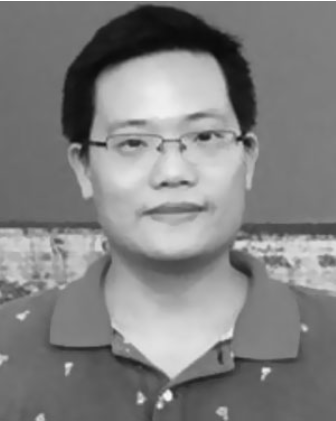}}]{Yongjian Wu}
   received his master’s degree in computer science from Wuhan University, China, in 2008. He is currently the expert researcher and the director of the Youtu Lab, Tencent Co., Ltd. His research interests include face recognition, image understanding, and large scale data processing.
\end{IEEEbiography}

\begin{IEEEbiography}[{\includegraphics[width=1in,height=1.25in,clip,keepaspectratio]{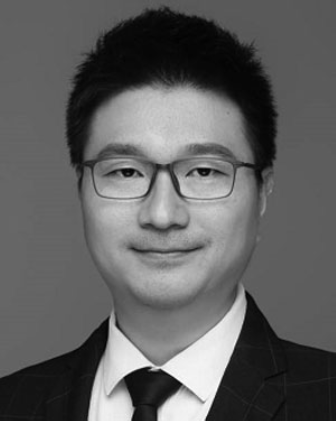}}]{Feiyue Huang}
   received his B.S. and Ph.D. degrees in computer science from Tsinghua University, China, in 2001 and 2008, respectively. He is the expert researcher and the director of the Tencent Youtu Lab. His research interests include image understanding and face recognition.
\end{IEEEbiography}

\begin{IEEEbiography}[{\includegraphics[width=1in,height=1.25in,clip,keepaspectratio]{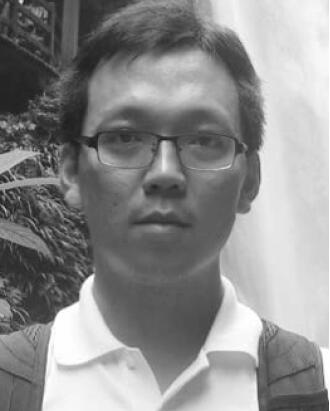}}]{Rongrong Ji}
(Senior Member, IEEE) is currently a
Professor and the Director of the Intelligent Multimedia
Technology Laboratory,  School of Informatics, Xiamen University, Xiamen, China.
His work mainly focuses on innovative technologies
for multimedia signal processing, computer
vision, and pattern recognition, with over 100 papers
published in international journals and conferences.
He is a member of the ACM. He also serves as a
program committee member for several Tier-1 international
conferences. He was a recipient of the ACM Multimedia Best Paper
Award and the Best Thesis Award of Harbin Institute of Technology. He serves
as an Associate/Guest Editor for international journals and magazines, such
as \textit{Neurocomputing},\textit{ Signal Processing}, \textit{Multimedia Tools and Applications},
the \textit{IEEE Multimedia Magazine}, and \textit{Multimedia Systems}.
\end{IEEEbiography}







\end{document}